\documentclass[conference]{IEEEtran}
\IEEEoverridecommandlockouts
\usepackage{cite}
\usepackage{amsmath,amssymb,amsfonts}
\usepackage{algorithmic}
\usepackage{graphicx}
\usepackage{textcomp}
\usepackage{xcolor}
\usepackage{subfig}
\usepackage[linesnumbered,ruled,vlined]{algorithm2e}
\def\BibTeX{{\rm B\kern-.05em{\sc i\kern-.025em b}\kern-.08em
    T\kern-.1667em\lower.7ex\hbox{E}\kern-.125emX}}

\begin{document}

\title{Cube Sampled $K$-Prototype Clustering for Featured Data}

\author{\IEEEauthorblockN{Seemandhar Jain}
\IEEEauthorblockA{\textit{Computer Science and Engineering} \\
\textit{Indian Institute of Technology Indore}\\
Indore, India \\
cse170001046@iiti.ac.in}
\and
\IEEEauthorblockN{Aditya A. Shastri}
\IEEEauthorblockA{\textit{Information Technology} \\
\textit{NMIMS, Shirpur}\\
Shirpur, India \\
aditya.shastri@nmims.edu}
\and
\IEEEauthorblockN{Kapil Ahuja}
\IEEEauthorblockA{\textit{Math of Data Science \& Simulation (MODSS) Lab} \\
\textit{Indian Institute of Technology Indore}\\
Indore, India \\
kahuja@iiti.ac.in}
\and
\IEEEauthorblockN{Yann Busnel}
\IEEEauthorblockA{\textit{Network Systems, Cybersecurity and Digital Law Department} \\
\textit{Institut Mines-Telecom Atlantique}\\
Rennes, France \\
yann.busnel@imt-atlantique.fr}
\and
\IEEEauthorblockN{Navneet Pratap Singh}
\IEEEauthorblockA{\textit{Computer Science and Engineering} \\
\textit{NMIMS Indore}\\
Indore, India \\
navneet.diat@gmail.com}

}

\maketitle

\begin{abstract}
Clustering large amount of data is becoming increasingly important in the current times. Due to the large sizes of data, clustering algorithm often take too much time. Sampling this data before clustering is commonly used to reduce this time. In this work, we propose a probabilistic sampling technique called cube sampling along with $K$-Prototype clustering. Cube sampling is used because of its accurate sample selection. $K$-Prototype is most frequently used clustering algorithm when the data is numerical as well as categorical (very common in today's time). The novelty of this work is in obtaining the crucial inclusion probabilities for cube sampling using Principal Component Analysis (PCA).

Experiments on multiple datasets from the UCI repository demonstrate that cube sampled $K$-Prototype algorithm gives the best clustering accuracy among similarly sampled other popular clustering algorithms ($K$-Means, Hierarchical Clustering (HC), Spectral Clustering (SC)). When compared with unsampled $K$-Prototype, $K$-Means, HC and SC, it still has the best accuracy with the added advantage of reduced computational complexity (due to reduced data size).  
\end{abstract}


\section{Introduction}
Due to the pervasiveness of social media and the subsequent transformation of the world digitally, an extensive amount of data is being generated. According to a survey \cite{milenkovic201930}, more than 2.5 quintillion bytes of data is produced every day, which is increasing exponentially. Since not all of this data is useful, it is important to obtain some meaning from it to help in decision making.

Clustering is the technique to group together the data items having similar behavior. It is often used to get the initial perception of data and is widely used.
Some of the popular clustering algorithms include $K$-Means, $K$-Modes, $K$-Prototype, Hierarchical Clustering (HC), Spectral Clustering (SC), Partitioning Around Medoids (PAM) etc. \cite{kmeans,clustering_survey,hc,sc}. 
$K$-Prototype is a commonly used clustering algorithm because of its ability to handle numerical and categorical data together \cite{huang1998extensions}. To achieve this, it uses a combination of the $K$-Means and the $K$-Modes algorithms and is easy to implement as well.   

Clustering becomes difficult when the size of the data is very large. One of the ways to develop an efficient clustering algorithm is to sample this data. There are several sampling techniques that are commonly used. This include random sampling, stratified sampling, Vector Quantization (VQ), pivotal sampling, cube sampling etc. \cite{random,shastri2019vector, shastri2020probabilistically,tille2006sampling}. We focus on cube sampling because {\it firstly} it belongs to the class of probabilistic sampling that are accurate. {\it Secondly}, it is independent of the order of the data selected to compute the crucial inclusion probabilities \cite{tille2006sampling}. To sum up, we propose cube sampled $K$-Prototype clustering algorithm with specific focus on data that contains numerical and categorical features.

There have been a few attempts to perform clustering while using a sampling technique to reduce the data \cite{shastri2019vector, shastri2020probabilistically}. In \cite{shastri2019vector}, authors have proposed use of VQ sampling technique along with SC. 
VQ is a very basic sampling technique with comparatively low accuracy. Moreover,  the algorithm in \cite{shastri2019vector} is specifically designed for genome data. Another work is \cite{shastri2020probabilistically}, where a different probabilistic sampling technique called pivotal sampling is used with SC. Although, pivotal sampling is good in terms of accuracy, this algorithm is tested on only plant phenotypic data.
This paper also uses pivotal sampling with HC. However, it also has same limitations as pivotal sampling with SC. In another work \cite{yang}, the authors have used random sampling with the ensemble clustering algorithm. However, they do not show the efficacy of their approach on large datasets. 

As evident from above discussion, cube sampling has many advantages and it has not been coupled with any of the commonly used clustering algorithms, which is the focus of this work. 
As mentioned above, besides combining cube sampling with $K$-Prototype, we integrate it with $K$-Means, HC, and SC as well.
The most novel contribution of this work is in computing the crucial inclusion probabilities for the cube sampling technique using the concept of Principal Component Analysis (PCA). 

The proposed algorithm is evaluated with extensive experiments that are performed on six labeled datasets from UCI repository with varying sizes.
Clustering Accuracy is used as a metric for comparison. 
We perform four sets of experiments. {\it First}, we compare four commonly used clustering algorithms (i.e. unsampled $K$-Prototype, $K$-Means \cite{kmeans}, HC \cite{hc}, and SC \cite{sc}), where the $K$-Prototype gives the best accuracy. {\it Second}, when using standard random sampling \cite{random}, we again compare these algorithms. Here, again, $K$-Prototype with random sampling gives the best accuracy.
{\it Third}, we perform same comparison with our cube sampling. {\it Fourth} and {\it finally}, we compare our cube sampled $K$-Prototype with unsampled other clustering algorithms mentioned above. 
Results for both the third and the fourth sets of comparisons show that our proposed algorithm performs the best. Also, out cube sampled $K$-Prototype has the added advantage of reduced computational complexity (since the input data is reduced by cube sampling before clustering).

The rest of this paper is organized as follows. 
Section \ref{sec:pa} provides a detailed discussion of our proposed algorithm.
Experimental setup is presented in Section \ref{setup}. 
Section \ref{sec:analysis} gives the experimental results. Finally, conclusions and future work are provided in Section \ref{sec:concl}.

\section{Proposed Algorithm}\label{sec:pa}
In this section, we first explain the concept of cube sampling including our novel way of obtaining the inclusion probabilities. Subsequently, we discuss the $K$-Prototype clustering algorithm, followed by our cube sampled $K$-Prototype clustering.  

\subsection{Application of Cube Sampling}
\label{cubesampling}
Cube sampling is a type of probabilistic sampling technique that selects a balanced sample \cite{tille2006sampling}.
Consider a population $U$ of size $n$ such that $U=\{x_{1},x_{2},..,x_{n}\}$, where $x_i \in \mathbb{R}^m$ for $i=1, 2, ..., n$ and $m$ is the total number of features. A sample $S$ of size $N$ is said to be balanced if the following equality is satisfied \cite{chauvet2006fast}.
\begin{equation}\label{eq1}
\sum_{j \in S} \frac{x_{j}}{\pi_{j}}=\sum_{i \in U} x_{i},
\end{equation}
where $\pi_j$ is the inclusion probability of the $j^{th}$ unit and $j=1, 2, ..., N$. Obtaining the inclusion probabilities of all the units in the population, denoted by $\pi_i$, forms an important aspect of this probability based sampling technique.

Cube sampling obtains the samples in two phases; the flight phase and the landing phase. The flight phase starts by obtaining the initial values of $\pi_i$ for $i=1, 2, ..., n$ (our unique contribution; discussed in the subsequent paragraphs). These initial probabilities are updated in each iteration of flight phase using a random walk. At each iteration, at least one unit is either included or excluded from the sample. At the end of this phase, inclusion probabilities of all the units are either equal to 0 (eliminated from the sample) or equal to 1 (included in the sample).

After the flight phase, if the balancing equation is not exactly satisfied or, inclusion probability of even one unit is neither 0 nor 1, then there is a need for the landing phase. The goal of the landing phase is to find a sample $S$ such that it is almost balanced (i.e. as close to balanced as possible).

Let $p$ be the number of units having inclusion probabilities neither 0 nor 1. Here, we obtain all the possible samples by taking the probability values of these $p$ units both 0 and 1. As there are two possible values for each of these $p$ units, we get a total of $2^p$ possible samples. Finally, we select the sample that closely satisfies the balancing condition. 

Next, the working of cube sampling is explained using an example of a random population of size six. Let the initial probabilities values are $\pi^{0}=(0.5,0.5,0.5,0.5,0.5,0.5)$, and we need to select three samples. The cube sampling transforms these initial inclusion probabilities as shown in (\ref{example}). Here, units with the final probability as 0 are discarded and units with the final probability as 1 are selected in the sample.
\begin{multline}\label{example}
\pi^{0}=\left(\begin{array}{c}
0.5 \\
0.5 \\
0.5 \\
0.5 \\
0.5 \\
0.5
\end{array}\right) \rightarrow\left(\begin{array}{c}
0.6 \\
0.6 \\
0 \\
0.6 \\
0.6 \\
0.6
\end{array}\right) \rightarrow\left(\begin{array}{c}
0.6 \\
0 \\
0 \\
1 \\
0.8 \\
0.6
\end{array}\right) \\ \rightarrow \left(\begin{array}{c}
1 \\
0 \\
0 \\
1 \\
0.5 \\
0.5
\end{array}\right) \rightarrow\left(\begin{array}{l}
1 \\
0 \\
0 \\
1 \\
1 \\
0
\end{array}\right)=\pi^{n}
\end{multline}

Inclusion probability of an unit $x_{i}$ is defined as the probability with which the unit is selected in the sample $S$ and, as earlier, is denoted by $\pi_{i}$. Here, we propose a novel and an efficient way to calculate the initial inclusion probabilities using Principle Component Analysis (PCA) \cite{karamizadeh2013overview}. 
Although PCA has been used earlier to sample data, it has never been combined with any probabilistic sampling techniques. 
PCA transforms the data by projecting it onto a single axis where elements are maximally deviated. The process of calculating the inclusion probabilities is described in the following equations \cite{tille2006sampling}:
\begin{equation}\label{eq4}
\pi_{i}=\min \left[1, h^{-1}(n) \frac{x_{i}}{X}\right], i \in U
\end{equation}
where, as earlier, $U$ is the total population and $h$ is defined as

\begin{equation}
h(z)=\sum_{i \in U} \min \left(z \frac{x_{i}}{X}, 1\right),
\end{equation}
with $\sum_{i \in U} \pi_{i}=N$ and $X=\sum_{i \in U} x_{i}$. 

\subsection{Our Algorithm}
The {$K$-Prototype} algorithm is an integration of the $K$-Means algorithm and the $K$-Modes algorithm to handle mixed data types. Here, the basic idea is similar to $K$-Means with a different distance criteria to handle numerical and categorical features. The aim is to minimize the cost function by partitioning the dataset into $k$ clusters. This cost function is given as 
\begin{equation}\label{eqc}
C=\sum_{l=1}^{k} \sum_{j=1}^{N} p_{jl} \text{DIST}\left(x_{j}, q_{l}\right),
\end{equation}
where $p_{jl} \in\{0,1\}$ denotes the membership of data point $x_{j}$ in cluster $l$, $q_{l}$ denotes the cluster center for cluster $l$ and DIST$\left(x_{j}, q_{l}\right)$ is the distance criteria (dissimilarity measure), which is given as
\begin{equation}\label{eqd}
\text{DIST}\left(x_{j}, q_{l}\right)=\sum_{r=1}^{m_{r}}\left(x_{jr}-q_{l r}\right)^{2}+\gamma_{l} \sum_{t=1}^{m_{t}} \delta\left(x_{j t}, q_{lt}\right),
\end{equation}
where, 
\begin{itemize}
	\item $r$ is index for numerical features and $m_r$ is the total number of numerical features for each data point $x_j$,
	\item $x_{ir}$ denotes the value of the $r^{th}$ numerical feature of $x_j$,
	\item $q_{lr}$ is the mean of the $r^{th}$ numerical feature in the cluster $l$,
	\item $\gamma_{l}$ is the weight factor for categorical features for cluster $l$,
	\item $t$ is index for categorical features and $m_t$ is the total number of categorical features of $x_j$,
	\item $x_{jt}$ denotes the $t^{th}$ categorical feature of $x_j$,
	\item $q_{lt}$ is the mode of the $t^{th}$ categorical feature in the cluster $l$, and
	\item $\delta(\mathrm{a}, \mathrm{b})=1$ for $a \neq b$ and $\delta(\mathrm{a}, \mathrm{b})=0$ for $a=b$. 
\end{itemize}

\begin{figure}[!h]
	\centering
	\includegraphics[width=\linewidth]{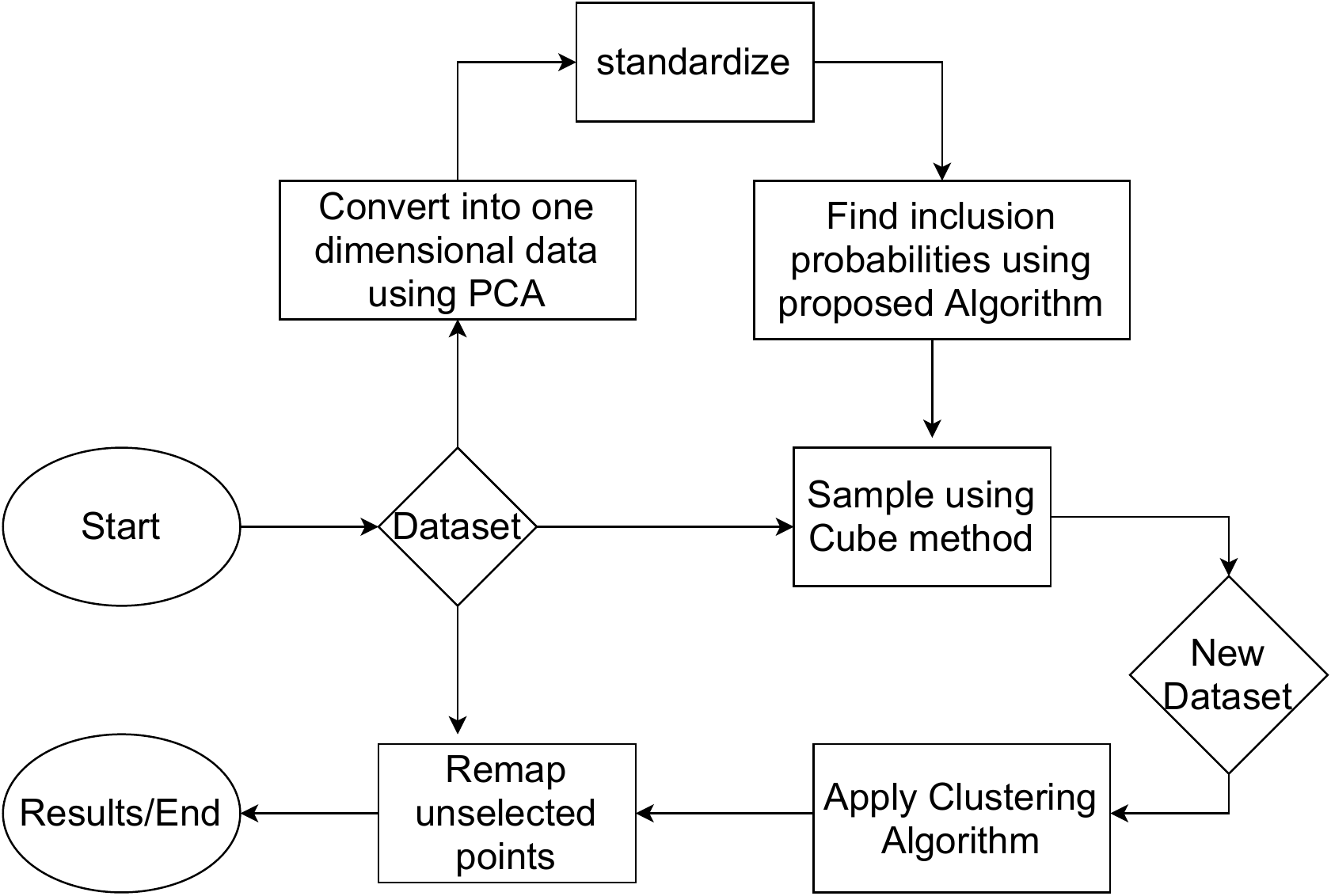}
	\caption{Framework for the proposed algorithm.}
	\label{fig:framework2}
\end{figure}

The proposed algorithm (Cube sampled $K$-Prototype algorithm) is given in Algorithm \ref{al4} and the broad framework is graphically represented in Fig. \ref{fig:framework2}.

\begin{algorithm}[!h]
	\DontPrintSemicolon
	\KwIn{$A$- Data Matrix ($n\times m$ rows), $N$- Sampling size, $k$- Number of clusters}
	\KwOut{Clusters}
	$\pi$ denotes initial probabilities that is computed as discussed in Section \ref{cubesampling}\;
	$\pi_{n}$ denotes final probabilities that is obtained after the flight and the landing phase of cube sampling (also discussed in Section \ref{cubesampling})\;
	
	Generate sample S from $\pi_{n}$ by using units with probability 1\;
	Choose $k$ cluster centers from $S$ randomly to form $q_{old}$\;
	Find the distance between elements and cluster centers as described in (\ref{eqd}) to form $q$.\;
	\While{$q_{old}\neq q$}{
		$q_{old} = q$\;
		Update $q$ using (\ref{eqd}) again\;
	}
	Reverse map the remaining points based on their similarity with all clusters.\;
	\Return{$Cluster Details$}
	\caption{Proposed Algorithm}
	\label{al4}
\end{algorithm}

\section{Experimental Setup}\label{setup}
In this section, we first briefly discuss the datasets used for our experiments and then describe the criteria used to check the goodness of our proposed algorithm.
For experiments, we use six datasets with varying sizes. These include labeled datasets downloaded from the UCI repository \cite{Dua:2019}. Table \ref{dataset} provides a brief summary of these datasets, where all the columns are self explanatory.

\begin{table}[!h]
	\centering
	\normalsize
	\caption{Dataset information from the UCI repository}
	\label{dataset}
	\begin{tabular}{|c|c|c|c|} 
		\hline
		\textbf{Dataset}   & \textbf{\# of records} & \textbf{\# of features} & {\bf \# of classes}  \\ 
		\hline
		German    & 1000              & 20                 & 2                  \\ 
		\hline
		Heart     & 303               & 13                 & 2                  \\ 
		\hline
		Satellite & 6435              & 36                 & 2                  \\ 
		\hline
		Adult     & 45222             & 14                 & 2                  \\ 
		\hline
		Shuttle   & 43500             & 9                  & 7                  \\ 
		\hline
		Kddcup99  & 24701             & 41                 & 18                 \\ 
		\hline
	\end{tabular}
	
\end{table}

Since we use labeled datasets for our experiments, we check the quality of our clustering by using Clustering Accuracy (CA), which is calculated as given below \cite{clustering_survey}.  
\begin{equation}\label{ca}
C A=\sum_{l=1}^{k} \frac{\max \left(C_{l} \mid L_{l}\right)}{|U|},
\end{equation}
\noindent
where $C_{l}$ is the set of elements in the $l^{th}$ cluster, $L_{l}$ is the set of all the class labels in the $l^{th}$ cluster, $max(C_{l}\mid L_{l})$ is the number of elements with the majority label in the $l^{th}$ cluster, and $U$ is the population size.

The experiments are performed on a machine with Intel® Core™ \textit{i7-7500U} CPU $2.90$ $GHz$ and $16$ $GB$ RAM. Python $3.7$ is used to implement all the algorithms.

\section{Results}\label{sec:analysis}
In this section, we show the effectiveness of our proposed algorithm (i.e. Cube sampled $K$-Prototype). 
For this, we do four various comparisons here. 
\begin{itemize}
	\item Without performing any sampling, we first compare $K$-Prototype clustering with other commonly used clustering techniques ($K$-Means, Hierarchical Clustering (HC), Spectral Clustering (SC)).
	\item When using frequently used random sampling, we again compare $K$-Prototype with same set of clustering algorithms. 
	\item Next, when using our proposed cube sampling, we further compare all the above-mentioned clustering algorithms.
	\item Finally, we compare our cube sampled $K$-Prototype algorithm with unsampled other clustering algorithms mentioned above.
\end{itemize}

Table \ref{clusterunsamp} compares four different clustering algorithms ($K$-Prototype, K-Means, Hierarchical Clustering (HC), and Spectral Clustering (SC)) without applying any prior sampling of data. From this table, it is clear that CA values for $K$-Prototype are greater than or equal to the other algorithms for all the datasets. Thus, $K$-Prototype performs best among all.

\begin{table}[!h]
	\centering
	\normalsize
	\caption{CA values for four different clustering algorithms on the six labeled datasets without prior application of sampling.}
	\label{clusterunsamp}
	\begin{tabular}{|c|c|c|c|c|} 
		\hline
		\textbf{Dataset}   & $K$-\textbf{Prototype} & $K$-\textbf{Means} & \textbf{HC} & \textbf{SC}  \\ 
		\hline
		German    & 0.68        & 0.53    & 0.62                    & 0.66                 \\ 
		\hline
		Heart     & 0.59        & 0.52    & 0.53                    & 0.52                 \\ 
		\hline
		Satellite & 0.79        & 0.71    & 0.72                    & 0.75                 \\ 
		\hline
		Adult     & 0.69        & 0.61    & 0.67                    & 0.68                 \\ 
		\hline
		Shuttle   & 0.82        & 0.82    & 0.82                    & 0.82                 \\ 
		\hline
		Kddcup99  & 0.69        & 0.59    & 0.69                    & 0.68                 \\
		\hline
	\end{tabular}
\end{table}

Next, in Table \ref{clusterunsampranodm}, we again compare the four clustering algorithms mentioned above but on sampled data (using random sampling). Without loss of generality, we take sample size as 100 for the two relatively smaller datasets (German and Heart), while sample size is taken as 1000 for the other datasets. From this table, we again observe that $K$-Prototype with random sampling performs best when compared with all other algorithms with random sampling. Use of random sampling deteriorate the CA values substantially, which is not the case with our sampling technique (discussed below).

\begin{table}[!h]
	\centering
	\normalsize
	\caption{CA values for four different clustering algorithms on the six labeled datasets with prior application of random sampling.}
	\label{clusterunsampranodm}
	\begin{tabular}{|c|c|c|c|c|c|}
		\hline
		\textbf{Dataset}   & \textbf{N} & $K${\bf -Proto.} & \textbf{$K$-Means} & \textbf{HC} & \textbf{SC} \\ \hline
		German    & 100          & 0.51        & 0.39    & 0.43                    & 0.45                \\ \hline
		Heart     & 100          & 0.48        & 0.41    & 0.39                    & 0.42                \\ \hline
		Satellite & 1000         & 0.63        & 0.53    & 0.55                    & 0.56                \\ \hline
		Adult     & 1000         & 0.60        & 0.54    & 0.55                    & 0.57                \\ \hline
		Shuttle   & 1000         & 0.69        & 0.68    & 0.68                    & 0.69                \\ \hline
		Kddcup99  & 1000         & 0.41        & 0.37    & 0.39                    & 0.40                \\ \hline
	\end{tabular}
\end{table}

Further, in Table \ref{clusterunsamp1}, we perform the same set of comparisons as in Table \ref{clusterunsampranodm} except using data obtained after applying cube sampling.
The sample sizes here are same as in the previous set of comparisons. We can observe from this table that the CA values for $K$-Prototype with cube sampling (our proposed algorithm) is greater than all the other algorithms with cube sampling. 

\begin{table}[!h]
	\centering
	\normalsize
	\caption{CA values for four different clustering algorithms on the six labeled datasets with cube sampling.}
	\label{clusterunsamp1}
	\begin{tabular}{|c|c|c|c|c|c|} 
		\hline
		\textbf{Dataset}   & \textbf{N} & \textbf{$K$-Proto.} & \textbf{$K$-Means} & \textbf{HC}   & \textbf{SC}    \\ 
		\hline
		German    & 100           & 0.68        & 0.52    & 0.60 & 0.64  \\ 
		\hline
		Heart     & 100           & 0.59        & 0.50    & 0.52 & 0.51  \\ 
		\hline
		Satellite & 1000          & 0.78        & 0.69    & 0.70 & 0.73  \\ 
		\hline
		Adult     & 1000          & 0.70        & 0.60    & 0.64 & 0.67  \\ 
		\hline
		Shuttle   & 1000          & 0.82        & 0.80    & 0.81 & 0.82  \\ 
		\hline
		Kddcup99  & 1000          & 0.69        & 0.56    & 0.66 & 0.67  \\
		\hline
	\end{tabular}
\end{table}

Finally, we compare our proposed algorithm (cube sampled $K$-Prototype) with the unsampled four clusterings as discussed before. For this, we pick the data from the relevant columns of Table \ref{clusterunsamp} and Table \ref{clusterunsamp1}. This set of comparisons is given in Table \ref{tab:CA} below. We observe from this table that our algorithm performs better than all others.

\begin{figure}[!h]
	\centering
	\subfloat[German Dataset]{\label{fig:kdd1}\includegraphics[height=3cm]{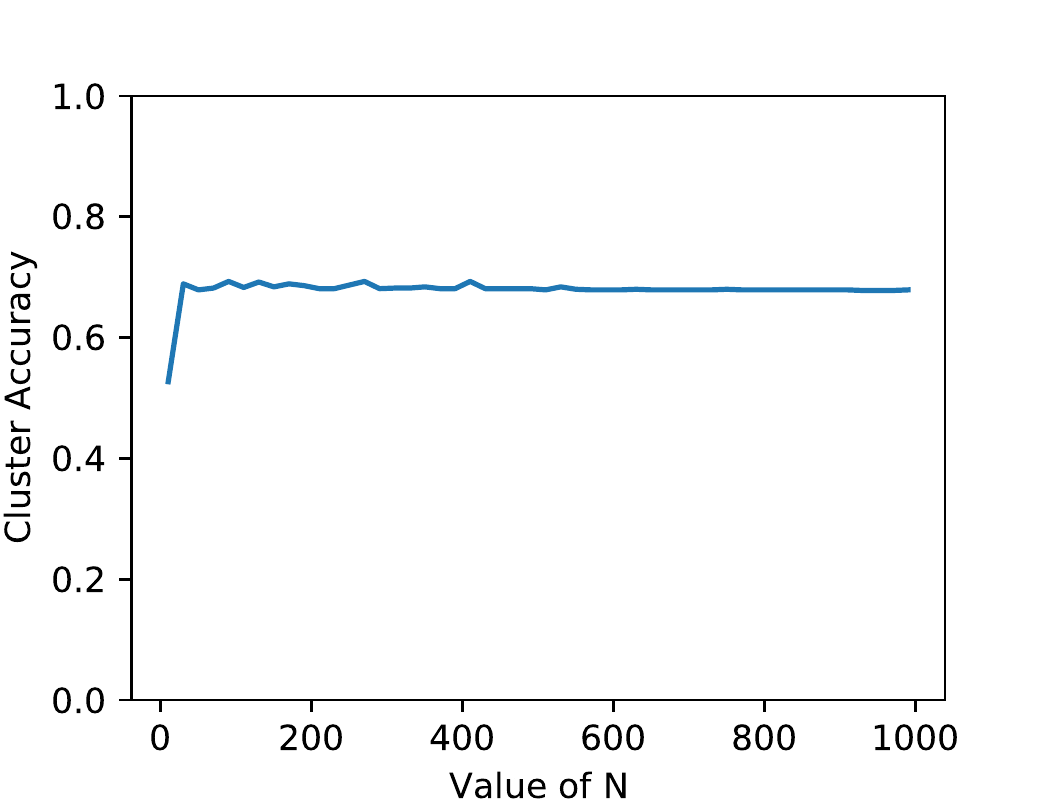}}
	\hfill
	\subfloat[Heart Dataset]{\label{fig:Satellite1}\includegraphics[height=3cm]{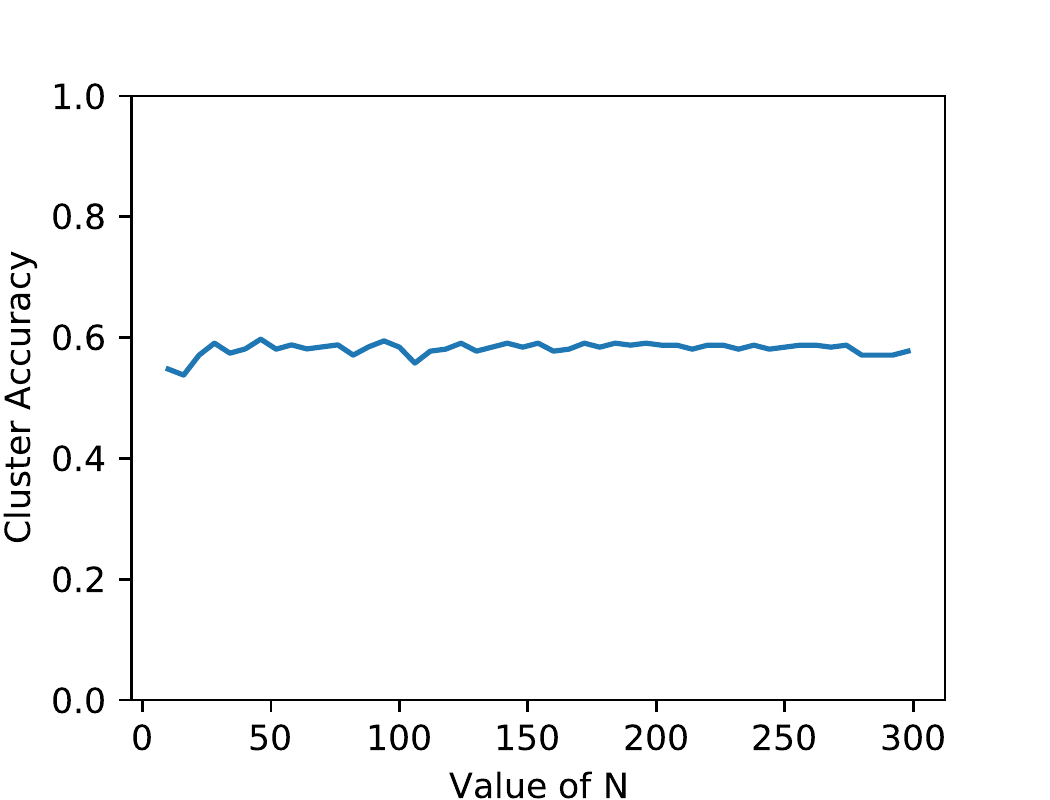}}
	\hfill
	\subfloat[Satellite Dataset]{\label{fig:Mul1}\includegraphics[height=3cm]{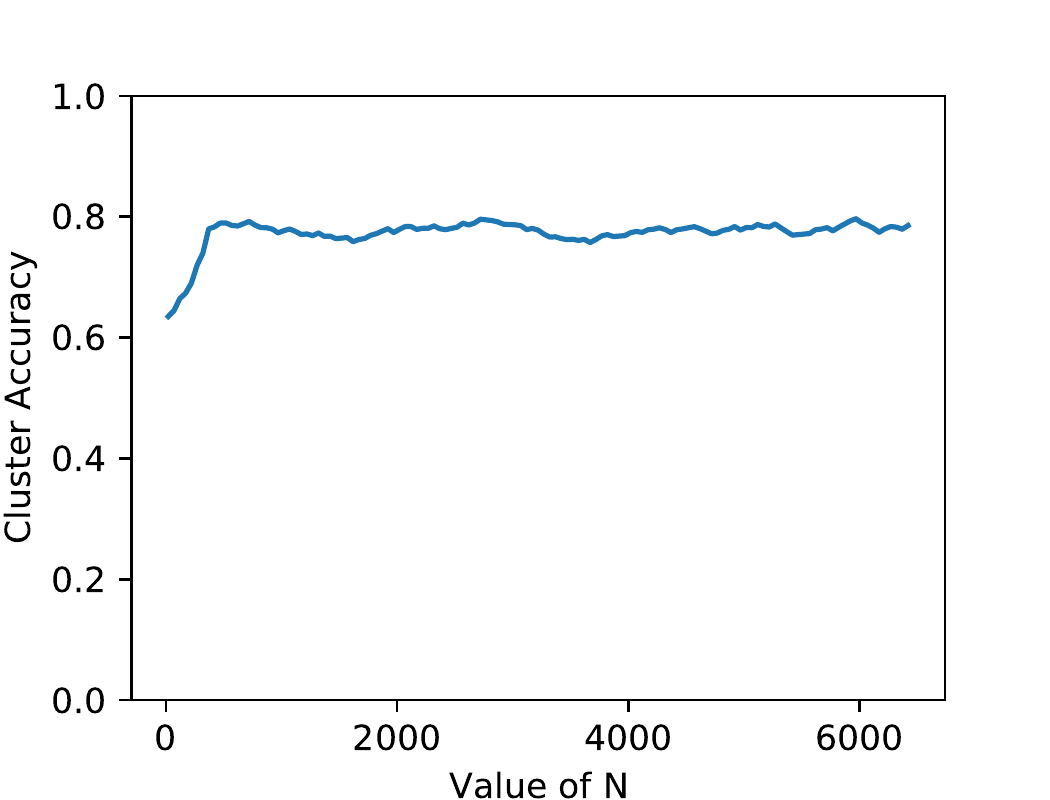}}
	\hfill
	\subfloat[Adult Dataset]{\label{fig:Forest1}\includegraphics[height=3cm]{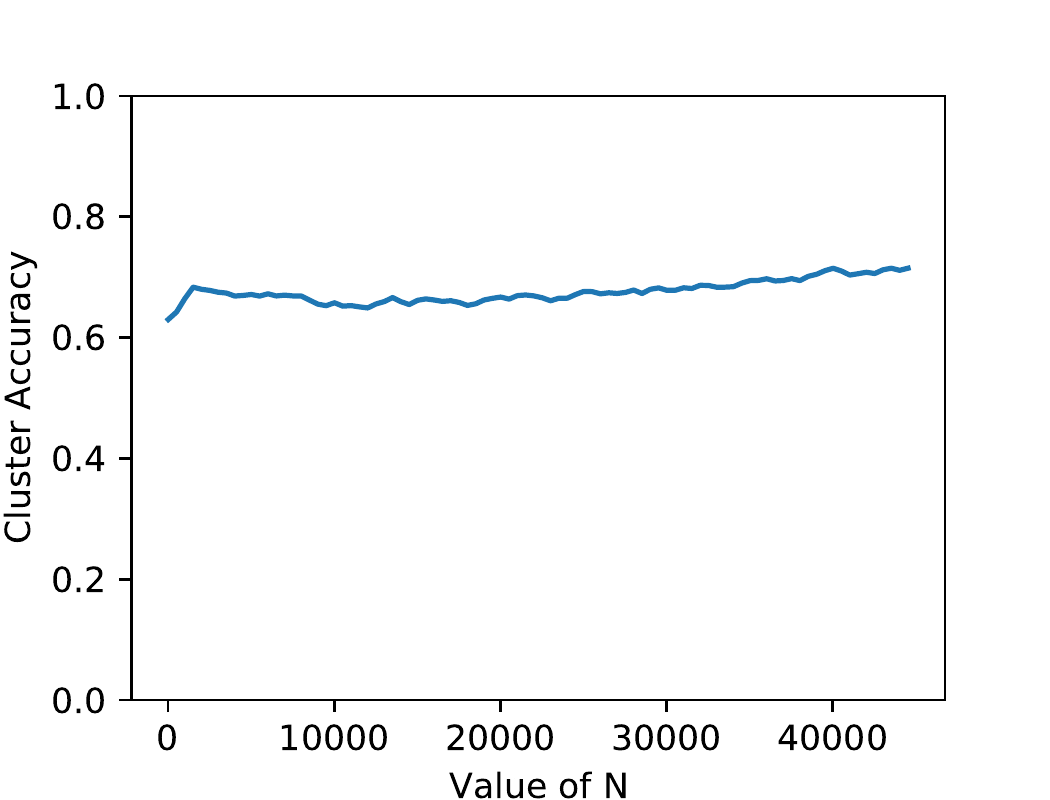}}
	\hfill
	\subfloat[Shuttle Dataset]{\label{fig:breast1}\includegraphics[height=3cm]{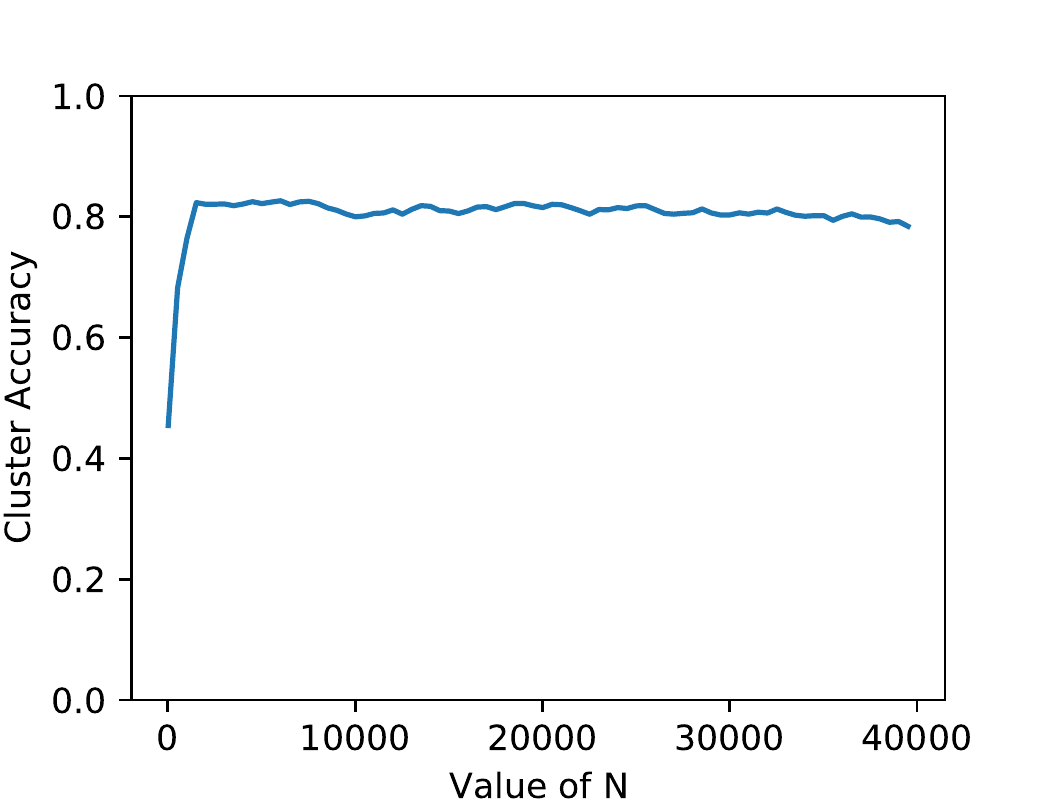}}
	\hfill
	\subfloat[Kddcup99 Dataset]{\label{fig:shuttle1}\includegraphics[height=3cm]{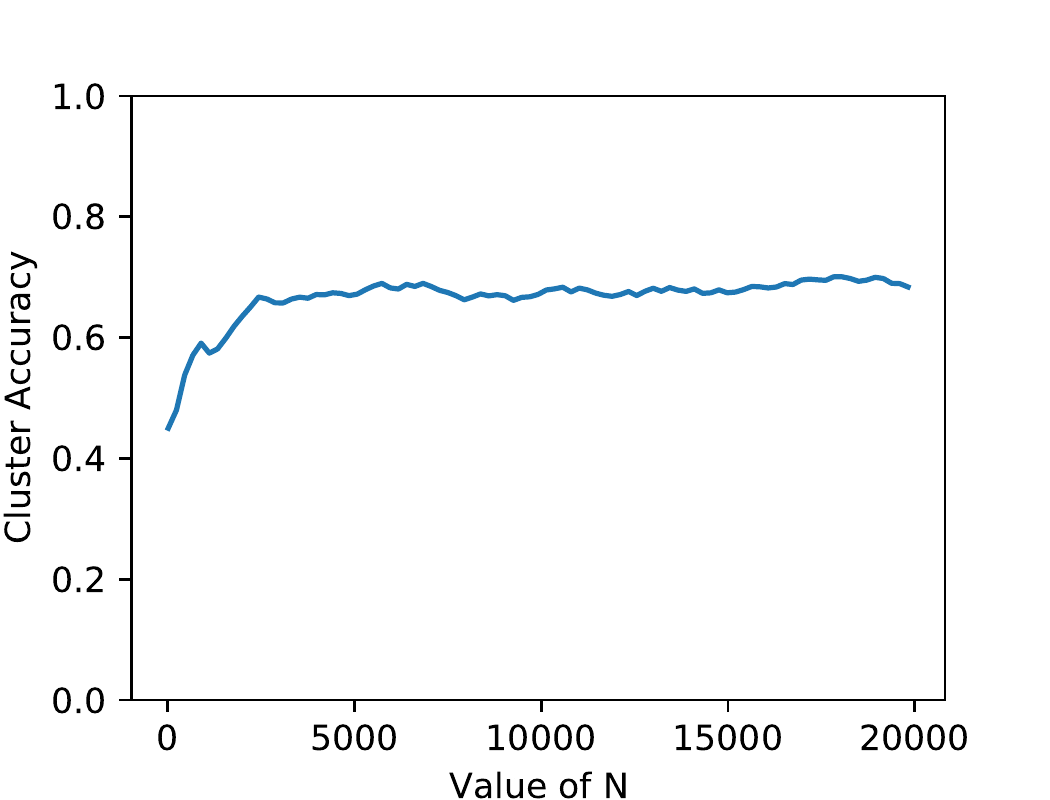}}
	\caption{CA values of our cube sampled $K$-Prototype algorithm for different sample sizes and different datasets.}
	\label{figure:Comp1} 
\end{figure}

To demonstrate that our technique gives good result irrespective of the size of the sample, in Fig. \ref{figure:Comp1}, we plot the CA values of our proposed algorithm for different sample sizes and different datasets. On the $x$-axis, sample size is given, and on the  $y$-axis, CA values are given. From these graphs, it is clear that sample size does not affect the CA values. Thus, use of cube sampling does not deteriorate the quality of clustering as done by $K$-Prototype. It also has an added advantage of reduced computational complexity (as it works on reduced dataset making it faster).  

 \begin{table}[!h]
 	\centering
 	\normalsize
 	\caption{CA values for our cube sampled $K$-Prototype algorithm and other four unsampled clustering algorithms on the six labeled datasets.}
 	\label{tab:CA}
 	\begin{tabular}{|p{1.3cm}|p{0.6cm}|p{0.6cm}|p{0.7cm}|p{0.9cm}|c|c|} 
 		\hline
 		\textbf{Dataset}   & $\boldsymbol{N}$ & {\bf Our Algo.} &
 		 $K$-\textbf{Proto.} & $K$-\textbf{Means} & \textbf{HC}  & \textbf{SC}   \\ 
 		\hline
 		German                   & 100  & 0.68                            & 0.68        & 0.53     & 0.62 & 0.66     \\ 
 		\hline
 		Heart                    & 100  & 0.59                            & 0.59        & 0.52     & 0.53  & 0.52    \\ 
 		\hline
 		Satellite                & 1000 & 0.78                            & 0.79        & 0.71     & 0.72   & 0.75   \\ 
 		\hline
 		Adult                    & 1000 & 0.70                            & 0.69        & 0.61     & 0.67  & 0.68    \\ 
 		\hline
 		Shuttle                  & 1000 & 0.82                            & 0.82        & 0.82     & 0.83  & 0.82    \\ 
 		\hline
 		Kddcup99                 & 1000 & 0.69                            & 0.69        & 0.59     & 0.69   & 0.68   \\
 		\hline
 	\end{tabular}
 \end{table}

\section{Conclusion}\label{sec:concl}

In this paper, we propose a cube sampled $K$-Prototype algorithm for grouping large data. 
The most innovative aspect of our algorithm is computation of inclusion probabilities in cube sampling using PCA. Experiments on labeled UCI datasets with varying sizes show that our proposed algorithm gives best clustering accuracy when compared with other commonly used unsampled clustering algorithms ($K$-Means, HC, and SC). Also, use of cube sampling does not deteriorate the accuracy of the $K$-Prototype algorithm. Furthermore, our algorithm also has the added advantage of reduced computational complexity (since the size of the data is reduced by cube sampling).
In future, we plan to adapt our algorithm for datasets with millions of units. We also aim to combine cube sampling with other accurate clustering algorithms. 
We intend to adapt our algorithm for Hadoop framework to make it parallel as well \cite{hadoop}.

%

\end{document}